\documentclass[letterpaper]{article} % DO NOT CHANGE THIS
\usepackage{aaai2026}  % DO NOT CHANGE THIS
\usepackage{times}  % DO NOT CHANGE THIS
\usepackage{helvet}  % DO NOT CHANGE THIS
\usepackage{courier}  % DO NOT CHANGE THIS
\usepackage[hyphens]{url}  % DO NOT CHANGE THIS
\usepackage{graphicx} % DO NOT CHANGE THIS
\usepackage{natbib}  % DO NOT CHANGE THIS AND DO NOT ADD ANY OPTIONS TO IT
\usepackage{caption} % DO NOT CHANGE THIS AND DO NOT ADD ANY OPTIONS TO IT
\usepackage{amsfonts}
\usepackage{amsmath} 
\usepackage{algorithm}
\usepackage{algorithmic}

\usepackage{amsfonts}
\usepackage{amsmath} 
\usepackage{booktabs}
\usepackage{colortbl}
\usepackage{makecell}
\usepackage{tabularx} 

\usepackage{tabularray}
\usepackage{array}

\usepackage{newfloat}
\usepackage{listings}
\DeclareCaptionStyle{ruled}{labelfont=normalfont,labelsep=colon,strut=off} % DO NOT CHANGE THIS
\floatstyle{ruled}
\newfloat{listing}{tb}{lst}{}
\floatname{listing}{Listing}
\title{Task-Aware 3D Affordance Segmentation via 2D Guidance and Geometric Refinement}
\author{
   Lian He\textsuperscript{\rm 1,\rm 2},
   Meng Liu\textsuperscript{\rm 2,\rm 3}\thanks{Corresponding Authors},
   Qilang Ye\textsuperscript{\rm 5,\rm 2},
   Yu Zhou\textsuperscript{\rm 5,\rm 2},
   Xiang Deng\textsuperscript{\rm 4},
   Gangyi Ding\textsuperscript{\rm 1*}
}
\affiliations{
    \textsuperscript{\rm 1} School of Computer Science \& Technology, Beijing Institute of Technology\\
    \textsuperscript{\rm 2} Zhongguancun Academy    \textsuperscript{\rm 3} Shandong Jianzhu University  \textsuperscript{\rm 4} Harbin Institute of Technology, Shenzhen\\
    \textsuperscript{\rm 5} VCIP \& TMCC \& DISSec, College of Computer Science \& College of Cryptology and Cyber Science, Nankai University\\
 
}

\usepackage{bibentry}
\begin{document}

\maketitle

\begin{abstract}
Understanding 3D scene-level affordances from natural language instructions is essential for enabling embodied agents to interact meaningfully in complex environments. However, this task remains challenging due to the need for semantic reasoning and spatial grounding. Existing methods mainly focus on object-level affordances or merely lift 2D predictions to 3D, neglecting rich geometric structure information in point clouds and incurring high computational costs. To address these limitations, we introduce \textbf{T}ask-\textbf{A}ware 3D \textbf{S}cene-level \textbf{A}ffordance segmentation (TASA), a novel geometry-optimized framework that jointly leverages 2D semantic cues and 3D geometric reasoning in a coarse-to-fine manner. To improve the affordance detection efficiency, TASA features a task-aware 2D affordance detection module to identify manipulable points from language and visual inputs, guiding the selection of task-relevant views. To fully exploit 3D geometric information, a 3D affordance refinement module is proposed to integrate 2D semantic priors with local 3D geometry, resulting in accurate and spatially coherent 3D affordance masks. Experiments on  SceneFun3D demonstrate that TASA significantly outperforms the baselines in both accuracy and efficiency in scene-level affordance segmentation. Projrct: https://hl1208649851.github.io/TASA/

\end{abstract}


\section{Introduction}
Affordances describe the latent functional properties of objects, such as handles, knobs, and switches, which allow specific actions~\cite{gibson2014ecological}. Understanding affordances in 3D environments is essential for enabling embodied agents to interact effectively with the physical world~\cite{c:1}. To perform goal-directed behaviors, agents must not only perceive these functional components but also ground them with high-level instructions, often expressed through natural language. This grounding is especially challenging in cluttered environments, where language instructions implicitly reference specific, functionally relevant object parts. For example, executing the command ``open the bottom drawer'' requires identifying and localizing the correct handle among many visually similar alternatives. Solving this task demands:
\textit{1) commonsense reasoning to infer implied functional targets from language}, and
\textit{2) fine-grained spatial perception to detect small and context-sensitive elements}. Effectively bridging natural language and 3D perception under such conditions is a critical step toward building intelligent agents capable of operating autonomously in real-world environments.

%Affordance understanding in 3D scenes requires interpreting natural language descriptions to locate functional interactive elements, such as handles, knobs, and buttons, in a 3D environment\cite{c:1}. Enabling embodied agents to interact seamlessly with the physical world hinges on their ability to understand affordance, the latent functional properties of objects that enable specific actions. In 3D scenes, this translates to segmenting fine-grained interactive elements such as light switches and drawer knobs from natural language task descriptions like ``turn on the ceiling light" or ``open the top drawer." This task is inherently challenging: it demands both world knowledge to interpret the implicit meaning of task descriptions and spatial perception to localize small, context-dependent functional objects within cluttered 3D environments such as distinguishing the bottom drawer handle from other handles in a cabinet.
\begin{figure}[t]
\centering
\includegraphics[width=1\columnwidth]{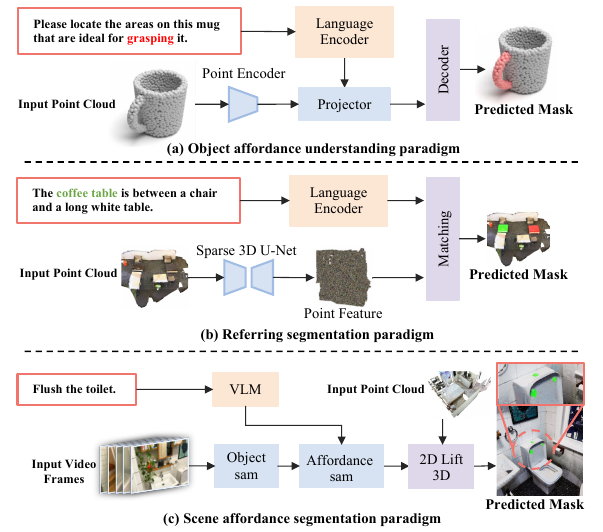} % Reduce the figure size so that it is slightly narrower than the column. Don't use precise values for figure width.This setup will avoid overfull boxes.
\caption{Comparison among (a) object affordance understanding, (b) referring segmentation, and (c) scene affordance segmentation paradigms.}
\label{fig1}
\end{figure}

Despite recent progress, existing approaches to 3D affordance understanding remain limited in several critical aspects. First, most methods adopt a narrow focus by targeting single-object affordance detection in isolation~\cite{3d-affordancellm, grounding3d, nguyen2023open, van2024open}. For example, they identify grasp points on a mug (Figure\ref{fig1}(a)) without considering the broader scene context. These object-centric paradigms, which treat individual objects as standalone inputs, \textbf{fail to scale to complex, scene-level reasoning} where affordances often depend on inter-object relationships and spatial arrangements. Second, many approaches rely heavily on 2D image-based information, typically projecting 2D segmentation masks into 3D via camera poses. This lifting process often overlooks essential geometric cues. As shown in Figure~\ref{fig1}(c), when handling the ``flush the toilet" task, it incorrectly segments point clouds with similar textures near the flush button into the affordance region, resulting in blurred segmentation outcomes that fail to differentiate between affordance-critical regions and surrounding areas with similar textures. \textbf{The absence of true 3D spatial constraints hampers the accurate localization of functional parts}. Third, current frameworks are inefficient in handling multi-view data. Typical methods for 3D affordance segmentation~\cite{Functionality} require processing hundreds of RGB frames, yet treat all views equally, \textbf{resulting in substantial redundancy and computational overhead without exploiting view-specific utility}.

To address the aforementioned challenges, we propose a \textbf{T}ask-\textbf{A}ware 3D \textbf{S}ece-level \textbf{A}ffordance segmentation framework (TASA), a novel geometry-optimized framework that jointly leverages 2D semantic cues and 3D geometric reasoning in a coarse-to-fine manner.
To our knowledge, this is the first geometry-optimized framework explicitly designed for affordance understanding in 3D scenes.

First, to move beyond the single-object paradigm, our method operates directly on full-scene 3D environments. It leverages vision-language models (VLMs) to parse task instructions and extract scene-level semantic cues from both 3D point clouds and multi-view RGB images. This enables context-aware affordance reasoning across objects, accounting for inter-object relationships and spatial configurations. 

Second, to reduce the inefficiency of uniform multi-view processing, we introduce a CLIP-based affordance-weighted frame selector. Guided by VLM-extracted task elements, it prioritizes views that are semantically aligned with the action, e.g., focusing on cabinet handles instead of irrelevant views like a ``TV'', for the instruction ``open the drawer of the cabinet with the TV on top''. This selective strategy significantly reduces redundancy while preserving task-relevant information.

Third, to mitigate over-reliance on 2D projections, we refine 2D segmentation masks by fusing them with 3D geometric features.
% enabling more accurate localization of functional parts in texture-similar regions. 
Given the complexity of directly interpreting high-resolution 3D point clouds, we use 2D masks as a guidance to partition the 3D point cloud, enabling the model to focus on critical regions and achieve precise localization of functional components in texture-similar areas.
In addition, we design a multi-objective loss function tailored for high-fidelity point clouds, where traditional methods often fail. This loss encourages fine-grained and spatially consistent segmentation across densely sampled scenes.

We evaluate our method on SceneFun3D, the only existing dataset with 3D affordance segmentation annotations in real-world indoor scenes. The dataset presents multiple challenges: 1) objects with similar visual appearances (e.g., dresser handles), 2) ambiguous and spatially complex task descriptions (e.g., ``open the second drawer''), and 3) dense, high-resolution point clouds that demand precise spatial reasoning. Our method demonstrates robust performance across all these scenarios.

Our contributions are summarized as follows:
\begin{itemize}
% \item We propose a Task-Aware 3D Scene-level Affordance segmentation framework, the first geometry-optimized framework designed for scene-level affordance segmentation.
% \item We propose a Task-Aware 3D Scene-level Affordance segmentation framework in a coarse-to-fine manner, integrating 2D cues with 3D geometric reasoning. 
\item We propose a Task-Aware 3D Scene-level Affordance segmentation framework that jointly leverages 2D semantic cues and 3D geometric reasoning in a coarse-to-fine manner.
\item We design a CLIP-based frame selection strategy guided by VLM-extracted task elements, and a 3D affordance refinement module to produce more precise and spatially consistent affordance masks.
% a CLIP-based affordance-weighted frame selector to prioritize key frames and a 3D affordance module to integrate geometry, significantly improving computational efficiency compared to previous methods that process per frame solely on 2D.
\item We achieve state-of-the-art results in 3D scene-level affordance segmentation with high-fidelity point clouds, significantly improving mAP from 7.6 to 23.2 and mIoU from 15.2 to 19.7.
\end{itemize}

\section{Related Work}
%We address the problem of 3D scene-level affordance segmentation, where the goal is to localize and segment fine-grained, task-relevant interactive components in complex environments. Given natural language instructions, multi-view video frames, and a 3D point cloud, our method grounds functional object parts, such as handles, knobs, or switches, to spatial regions actionable by an embodied agent. In this section, we review relevant work across three key areas: affordance understanding, 3D reference segmentation, and VLMs.

\begin{figure*}[t]
\centering
\includegraphics[width=1\textwidth]{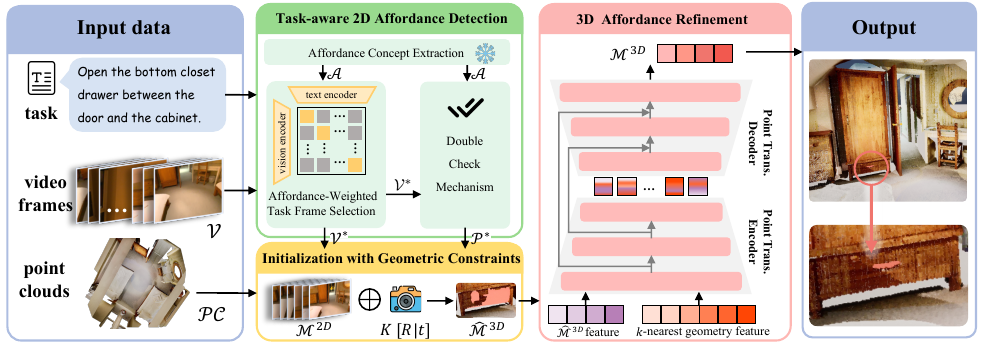} % Reduce the figure size so that it is slightly narrower than the column. Don't use precise values for figure width.This setup will avoid overfull boxes.
\caption{Overview of our Task-Aware 3D Scene-level Affordance segmentation framework (TASA).
%, which combines 2D semantic guidance with 3D geometric refinement in a coarse-to-fine pipeline.
}

\label{fig3}
\end{figure*}

\paragraph{Affordance Understanding.}
Affordance research spans both 2D and 3D domains. In 2D, recent efforts explore off-the-shelf segmentation models and weakly supervised learning~\cite{xu2024weakly,li2024one,xu2025weakly,qian2024affordancellm} for efficient RGB-based affordance parsing. However, these approaches lack 3D grounding, limiting their applicability to real-world embodied interaction.

In the 3D domain, most existing methods focus on object-centric affordance prediction~\cite{qian2024affordancellm,3d-affordancellm,grounding3d,lu2025geal,shao2025great,yang2024egochoir}. While some recent work attempts open-vocabulary affordance reasoning~\cite{nguyen2023open,van2024open}, these methods rely on isolated-object datasets and struggle to scale to complex scenes with clutter and occlusion. Their limited spatial context restricts fine-grained localization of functional parts in realistic settings. To our knowledge, Fun3DU~\cite{Functionality} is the only method targeting scene-level 3D affordance segmentation. However, it suffers from severe computational inefficiency and merely lifts 2D segmentations into 3D. This leads to cascading errors when initial 2D predictions are inaccurate.

In contrast, our method introduces a task-aware frame selection module that leverages CLIP embeddings guided by VLM-derived affordance cues. This drastically improves efficiency, reducing computation time by orders of magnitude compared to Fun3DU. Crucially, we further enhance segmentation robustness by integrating 3D geometric features into the refinement process, enabling reliable affordance localization in cluttered, high-fidelity environments.

\paragraph{3D Referring Segmentation.}
Among 3D scene understanding tasks, 3D referring segmentation is partially related to our problem setting. Given a natural language expression, these methods aim to segment the corresponding object in a 3D scene. Existing approaches fall into two main categories: referring expression-based methods~\cite{chen2020scanrefer,achlioptas2020referit3d,luo20223d}, and superpoint-based segmentation~\cite{wu20243d,sun2023superpoint,engelmann20203d}. While effective for whole-object localization, these methods lack the granularity needed for actionable manipulation. For example, the instruction ``close the bedroom door'' requires localizing the doorknob, not the entire door. Furthermore, most models are trained on coarse, low-resolution point clouds, limiting their generalization to high-fidelity 3D data common in real-world robotics applications.

\paragraph{Vision-Language Models.}
VLMs enable cross-modal reasoning through joint encoding of visual and textual inputs, and can be broadly categorized into 2D and 3D variants. 2D VLMs~\cite{chen2024spatialvlm,c:1,li2022blip,alayrac2022flamingo,liu2018attentive,liu2018cross,ye2024pose,ye2024cat,ye2025cat+}, have seen rapid progress, driven by large-scale image-text pretraining. These models typically combine a vision encoder with a language model via various fusion architectures. For example, LLaVA~\cite{liu2023visual} fuses CLIP~\cite{c:1} features into a Vicuna-based language model using a lightweight MLP adapter. In contrast, 3D VLMs are still in their early stages, mainly focusing on grounding generic objects in 3D scenes. Due to the lack of large-scale 3D vision-language datasets and the absence of strong 3D foundation models, current 3D VLMs exhibit limited generalization, particularly in fine-grained affordance reasoning within complex environments.

\section{Method}
\subsection{Overview of the Framework}
Figure~\ref{fig3} illustrates our proposed framework for Task-Aware 3D Scene-level Affordance Segmentation, which adopts a coarse-to-fine pipeline to progressively improve segmentation accuracy. The objective is to localize actionable regions in a 3D point cloud based on natural language instructions.

Given a task description $\mathcal{T}$, a high-precision point cloud $\mathcal{PC} = \{p_i \in \mathbb{R}^3\mid i = 1, ..., M\}$, and set of multi-view RGB images $\mathcal{V} = \{v_1, v_2, ..., v_N\} \in \mathbb{R}^{N \times H \times W \times 3}$, where \(p_i\) is individual 3D points, $M$ denotes the total number of points in the point cloud, \(v_i\) refers to every RGB image, $N$ is the number of views, $H$ and $W$ are the image height and width, the framework outputs a 3D affordance mask indicating task-relevant regions suitable for downstream robotic manipulation. 

Our method comprises two main stages: 1) Task-aware affordance detection, which employs vision-language reasoning to guide informative frame selection, and  
2) 3D affordance mask refinement, which incorporates geometric features to ensure precise and spatially grounded segmentation.

\subsection{Task-Aware 2D Affordance Detection}
This module extracts reliable 2D manipulable points from natural language instructions and image frames to provide task-aligned guidance for downstream 3D segmentation. It consists of three components: 1) Affordance Concept Extraction using a VLM to identify task-relevant interactive components from text, 2) Affordance-Weighted Frame Selection via a CLIP-based similarity metric modulated by extracted concepts to prioritize informative views, and 3) Manipulable Point Validation through a double-check mechanism that enhances robustness via reverse checks and fallback queries.

\subsubsection{Affordance Concept Extraction.}

Given a natural language task $T$, we extract a set of interactive concepts essential for completing the instruction.  These typically correspond to manipulable object parts (e.g., ``handle'' in ``open the bottom closet drawer between the door and the cabinet''). Formally, let $\mathcal{A} = \{a_1, a_2, ..., a_k\}$ denote the set of extracted affordance concepts. These are obtained via the VLM: 
%denote the set of extracted affordance-related tokens, where each $a_i$ corresponds to a functional component (e.g., ``handle'' in ``open the bottom closet drawer between the door and the cabinet''). This extraction process is formalized as:
\begin{equation}
\mathcal{A}=\mathrm{VLM}_{\text{affordance}}(T;\theta_{\mathrm{VLM}}),
\end{equation}
where $\theta_{\mathrm{VLM}}$ represents the parameters of the VLM. The output $\mathcal{A}$ encodes functional cues semantically grounded in the instruction.

\subsubsection{Affordance-Weighted Frame Selection.}
Multi-view image streams often contain frames irrelevant to the task (e.g., occluded views, distant background clutter), which can degrade downstream performance and increase computational overhead. To address this, we propose a semantic frame selection strategy that emphasizes frames aligned with extracted affordance concepts. 

Standard CLIP-based similarity~\cite{c:1} tends to bias toward dominant nouns (e.g., ``cabinet''), neglecting functional components like ``handle''.  To mitigate this, we introduce an affordance-weighted similarity metric. Let $\mathrm{CLIP}_{\text{img}}: \mathbb{R}^{H \times W \times 3} \rightarrow \mathbb{R}^d$ and $\mathrm{CLIP}_{\text{txt}}: \mathcal{L} \rightarrow \mathbb{R}^d$ be the CLIP encoders for images and text, respectively, with embedding dimension $d = 512$. 
For each image frame $v_i \in \mathcal{V}$ and task $T$, we compute $\mathbf{e}_i^{\mathrm{img}}=\mathrm{CLIP}_{\mathrm{img}}(v_i)$, serving as  the reference embedding $\mathbf{e}^{\text{ref}}$ of frame $v_i$, $\quad\mathbf{e}_T^{\mathrm{txt}}=\mathrm{CLIP}_{\mathrm{txt}}(T)$, and the baseline CLIP similarity:
\begin{equation}
\mathrm{Sim}_{\mathrm{CLIP}}(v_i,T)=\frac{\mathbf{e}_i^\mathrm{img}\cdot\mathbf{e}_T^\mathrm{txt}}{\|\mathbf{e}_i^\mathrm{img}\|_2\cdot\|\mathbf{e}_T^\mathrm{txt}\|_2}\in[-1,1],
\end{equation}
serving as a baseline metric quantifying the overall semantic alignment between frame $v_i$ and task $\mathcal{T}$, capturing contextual information such as dominant entities (e.g., ``cabinet'') that provide scene-level relevance. To emphasize actionable components, we compute CLIP embeddings for each affordance concept $a_i \in \mathcal{A}$:
\begin{equation}
\mathbf{e}_{a_i}^{\mathrm{txt}}=\mathrm{CLIP}_{\mathrm{txt}}(a_i)\in\mathbb{R}^d.
\end{equation}
which are directly utilized in the affordance-weighted calculation. The affordance-weighted similarity score $S$, computed for each frame $v_i$:
\begin{equation}
\begin{cases}
S = \frac{1}{Z}\sum_{a_i \in \mathcal{A}} \exp(\mathbf{e}_{a_i}^{\text{txt}} \cdot \mathbf{e}^{\text{ref}}),\\
Z=\sum_{a_i\in\mathcal{A}}\exp(\mathbf{e}_{a_i}^{\mathrm{txt}}\cdot\mathbf{e}^{\mathrm{ref}}),
\end{cases}
\end{equation}
$\mathrm{Sim}_{\mathrm{CLIP}}(v_i,T)$ and $S$ are complementary and jointly contribute to frame selection. While $S$ prioritizes functional affordance concepts (e.g., handle) critical for task execution. $\mathrm{Sim}_{\mathrm{CLIP}}(v_i,T)$ ensures alignment with the broader task context (e.g., `cabinet''), preventing selection of irrelevant frames containing isolated affordance components.

For final frame selection, we caculate $\mathrm{FinalScore}(v_i)$ as:
\begin{equation}
\mathrm{FinalScore}(v_i) = \alpha_a \cdot \mathrm{Sim}_{\mathrm{CLIP}}(v_i,T) + (1-\alpha_a) \cdot S
\end{equation}
where $\alpha\in[0,1]$ balances contextual and affordance priorities. We rank frames by $\mathrm{FinalScore}(v_i)$ and select the top-$K$ frames as $\mathcal{V}^*$, prioritizing semantically aligned, interactive elements over visually dominant. This method reduces the number of frames processed by focusing on the most relevant ones, enhancing both efficiency and accuracy.

%We enhance this by incorporating affordance concepts $\mathcal{A}$ generated by the VLM. For each $a\in\mathcal{A}$, compute its CLIP embedding $\mathbf{e}_a^{\mathrm{txt}}=\mathrm{CLIP}_{\mathrm{txt}}(a)\in\mathbb{R}^d$. The affordance-weighted similarity $S$ between a reference embedding $\mathbf{e}^{\text{ref}}$ and the affordance embeddings is defined in two steps. First, a normalization factor $Z$ is introduced as $Z=\sum_{a\in\mathcal{A}}\exp(\mathbf{e}_a^{\mathrm{txt}}\cdot\mathbf{e}^{\mathrm{ref}})$. With the normalization factor $Z$ in place, the affordance-weighted similarity $S$ can then be computed as: 
%\[S = \frac{1}{Z}\sum_{a \in \mathcal{A}} \exp(\mathbf{e}_a^{\text{txt}} \cdot \mathbf{e}^{\text{ref}}) \cdot w_a\] 
%where $w_a$ represents the weight associated with each affordance $a$, and $\mathcal{A}$ is the set of all affordances. This final formula fully captures the weighted similarity concept, ensuring that the influence of each affordance on the overall similarity measure is appropriately accounted for.

\subsubsection{Manipulable Point Validation.}
For each select frame $v \in \mathcal{V}^*$, we extract 2D manipulable points using a VLM, enhanced by a double-check mechanism to improve both robustness and accuracy.

We first issue a task-specific prompt to the VLM, asking it to identify spatial locations within the image that are critical for completing the given instruction  $T$. This yields a set of candidate points: $\mathcal{P}=\{(x_1,y_1),...,(x_m,y_m)\}$, which represent likely affordance regions (e.g., handles or switches) in the task-relevant frames.

\begin{figure}[t]
\centering
\includegraphics[width=1\columnwidth]{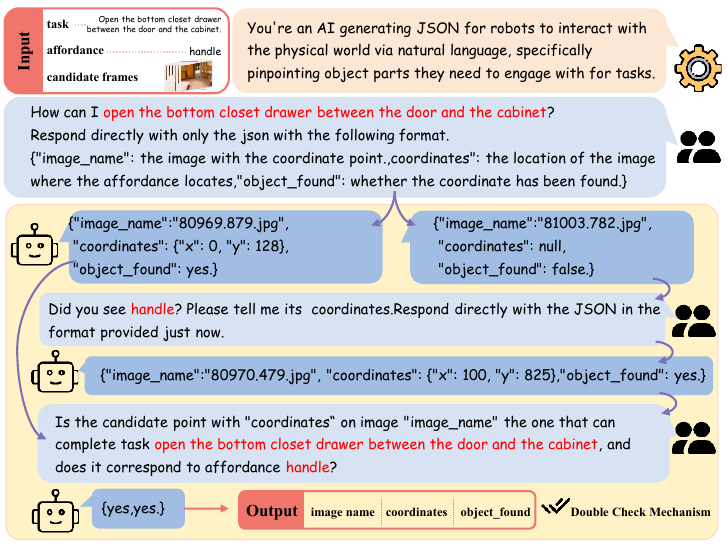} % Reduce the figure size so that it is slightly narrower than the column. Don't use precise values for figure width.This setup will avoid overfull boxes.
\caption{Illustration of the Double-Check Mechanism. 
%This module verifies whether the detected coordinates in candidate frames correspond to the task-relevant affordance components.
}
\label{fig4}
\end{figure}

However, direct one-shot VLM extraction often suffers from missed detections or false positives due to semantic ambiguity (e.g., the term ``handle'' may refer to visually diverse components). To address this, we introduce a double-check mechanism composed of two sequential steps: reverse validation and fallback querying, as illustrated in Figure~\ref{fig4}. 
\begin{itemize}
    \item \textbf{Reverse Validation}: For each candidate point $p_i^{2D} = (x_i, y_i) \in \mathcal{P}$, we verify whether interacting with this point is sufficient to complete task $T$. We define a verification function: $VLM_{verify}(v_i,p_i^{2D},T)$, which checks whether point $p_i$ corresponds to a functionally correct component (e.g., ensuring the point lies on a drawer handle rather than on the drawer surface). Points that fail validation are discarded, resulting a validation set of points, denoted as $\mathcal{P}^{valid}$
   \item \textbf{Fallback Query}: If no candidate point is extracted ($\mathcal{P} = \emptyset$), we trigger a fallback mechanism. For each affordance concept $a_i \in \mathcal{A}$ identified earlier, we query the VLM to estimate relevant 2D locations: $VLM_{point}(v_i,a_i;\theta_{VLM})$. This yields a supplementary set of points, denoted as $\mathcal{P}^*$.
\end{itemize}
The final set of verified manipulable points is the union of the reverse and fallback results (i.e., $\mathcal{P}^{valid} \cup \mathcal{P}^*$), ensuring both precision and recall in affordance localization. These points serve as coarse spatial cues for subsequent 2D mask generation, forming the initial step in our coarse-to-fine affordance segmentation pipeline.

%The double check mechanism consists of two sequential processes, as shown in Figure~\ref{fig4}~. In the reverse validation step, for each point $p = (x, y)$ within the set $\mathcal{P}$, we assess its sufficiency for task completion. This evaluation is formalized by applying a verification function $\text{VLM}_{\text{verify}}(v, p, T)$, where $v$ represents the visual input, and $T$ is the target task. In the fallback query step, where the set $\mathcal{P}$ is an empty set, an alternative computational pipeline is activated. For each element $a \in \mathcal{A}$, where $\mathcal{A}$ represents the object set, the $\text{VLM}_{\text{point}}(v, a; \theta_{\text{VLM}})$ function is employed to estimate the spatial coordinates of $a$  within the visual input $v$. The parameter $\theta_{\text{VLM}}$ is the parameter set of the visual language model. The resulting set of points, denoted as $\mathcal{P}^*$, encapsulates the newly detected spatial locations, effectively augmenting the initial point set. The final set of valid manipulable points is derived through a combination of these two processes, ensuring comprehensive coverage of potential task-completion locations.

\subsection{3D Affordance Refinement}
This module fuses 2D semantic guidance with 3D geometric features to generate accurate and spatially consistent 3D affordance masks. It consists of two key components: 1) Geometric Initialization, which projects 2D masks into 3D to produce coarse 3D mask candidates; and 2) a 3D Affordance Refinement module, which leverages local geometric context and self-attention to refine the masks.

\subsubsection{Geometric Initialization.}
Given selected task-relevant frames $\mathcal{V}^{*}$ and validated manipulable points $\mathcal{P}^{valid} \cup \mathcal{P}^*$, we first generate 2D binary masks for each frame $v_i\in\mathcal{V}^*$, denote as $\{\mathcal{M}^{2D}_v\}$ where $\mathcal{M}^{2D}_v\subseteq\{0,1\}^{H\times W}$ using a pre-trained segmentation model. These masks are then projected into the 3D point cloud $\mathcal{PC}=\{p_i\in\mathbb{R}^3\}_{i=1}^M$ using known camera intrinsics $K\in\mathbb{R}^{3\times3}$ and extrinsics $[R\mid t]\in\mathbb{R}^{3\times4}$, resulting in an initial 3D mask $\mathcal{M}_{\text{init}} \in \mathbb{R}^{M\times1}$. This provides a coarse binary indicator for each 3D point's relevance. 

To tackle the challenges of processing high-fidelity scene point clouds due to their scale and precision, we focus on points within $N$-neighbors of those points masked by \(\mathcal{M}_{\text{init}}\) to obtain \(\hat{\mathcal{M}}^{3D}\). This design aligns with our coarse-to-fine framework.

\subsubsection{3D Affordance Refinement.}
To capture local geometric dependencies in 3D space, we adopt the Point Transformer Layer inspired by ~\cite{zhao2021point}. This layer performs self-attention among neighboring points, enhanced by relative positional encoding. Given input point coordinates $p \in \mathbb{R}^{N \times 3}$ and features $x \in \mathbb{R}^{N \times C}$, the process is as follows:

First, the input features are projected into query, key, and value spaces via linear layers:
\begin{equation}
q_i = \text{W}_q x_i, k_j = \text{W}_k x_j, v_j = \text{W}_v x_j.
\end{equation}

For each point $p_i$, we identify its $k$ nearest neighbors $\{p_j\}_{j=1}^k$ and compute the relative positions $\delta_{i,j} = p_j - p_i$. These offsets are encoded by a multi-layer perceptron (MLP) to yield position embeddings:
\begin{equation}
\gamma_{i,j} = \text{MLP}_{\text{pos}}(p_j - p_i).
\end{equation}

Then, we compute attention weights using a combination of feature difference and position encoding. Specifically, for each point pair $(i, j)$, the unnormalized attention weight is computed as:
\begin{equation}
w_{i,j} = \text{MLP}_w\left( k_j - q_i + \gamma_{i,j} \right),
\end{equation}
where $\text{MLP}_w$ is a shared-weight network applied across local neighbors. These weights are normalized via a softmax function:
\begin{equation}
\alpha_{i,j} = \frac{\exp(w_{i,j})}{\sum_{l=1}^k \exp(w_{i,l})}.
\end{equation}

The output feature at each point is computed by aggregating the value features from its neighbors, weighted by attention and enriched with positional encoding:
\begin{equation}
x'_i = \sum_{j=1}^k \alpha_{i,j} \cdot (v_j + \gamma_{i,j}).
\end{equation}

Leveraging the Point Transformer Layer, we employ an encoder-decoder architecture for refining 3D affordance masks, as showed in Figure~\ref{fig3}.
The encoder aggregates local geometric features and downsamples the point cloud to capture global contextual information.
The decoder upsamples features via feature propagation layers with distance-weighted interpolation, while integrating skip connections to fuse encoder-derived features and retain fine-grained details, thereby ensuring precise segmentation boundaries.
% To improve efficiency, we apply channel grouping with a shared group size $s$, so that intermediate operations are performed in grouped feature space. 
% The layer can be stacked multiple times to hierarchically propagate geometric relations in the point cloud.

% With the feature capture Layer, we adopt an encoder-decoder architecture for point mask refine. 
% The encoder employs to aggregate local geometric features, downsampling the point cloud via farthest point sampling (FPS) to capture global context, transforming raw point coordinates and initial features into rich latent representations. 
% The decoder upsamples features through feature propagation layers using distance-weighted interpolation, with skip connections fusing encoder features to preserve fine-grained details for precise segmentation boundaries. 
% Final per-point features are processed by a multi-layer perceptron (MLP) and a softmax layer to predict semantic labels or affordance masks. Residual connections and layer normalization enhance training stability and convergence.

\subsubsection{Training Objective.}
To handle the challenges of mask prediction in high-resolution point clouds, we adopt a multi-objective loss function to supervise the training of the point transformer-based 3D refinement module. The total loss is defined as:
\begin{equation}
L_{\text{total}} = \alpha_1L_{\text{BCE}} + \alpha_2L_{\text{Dice}} + \alpha_3L_{\text{Focal}} + \alpha_4L_{\text{IoU}},
\label{loss}
\end{equation}
where  $\alpha_i \geq 0$ and $\sum_{i=1}^{4}\alpha_i = 1$. Each loss term addresses a distinct aspect of the 3D segmentation quality: 1) $L_{\text{BCE}}$ supervises per-point classification by minimizing pointwise prediction error. 2) $L_{\text{Dice}}$ encourages overlap between predicted and ground-truth regions, promoting region-level consistency. 3) $L_{\text{Focal}}$ emphasizes hard examples, mitigating class imbalance by down-weighting easy negatives. 4) $L_{\text{IoU}}$ directly optimizes the intersection-over-union between predicted and ground-truth masks, enhancing spatial alignment. This composite loss balances point-level precision and global spatial coherence, enabling the model to produce accurate and robust affordance masks in dense 3D scenes by jointly leveraging 2D semantic priors and 3D geometric context.

\begin{table}[t]
\centering
%\setlength{\tabcolsep}{2pt} 
%\footnotesize  
\begin{tabular}{ccccc} 
\toprule 
{Method} & {$\mathrm{mAP}$} & {$\mathrm{AP}_{50}$} & {$\mathrm{AP}_{25}$} & {$\mathrm{mIoU}$}\\ 
\midrule 
OpenMask3D & 0.2 & 0.2 & 0.4 & 0.2 \\ 
LERF & 0.0 & 0.0 & 0.0  & 0.0 \\ 
OpenIns3D & 0.0 & 0.0 & 0.0 & 0.1\\
Fun3DU & 7.6 & 16.9 & \textbf{33.3} & 15.2 \\
\midrule 
Ours & \textbf{23.2} & \textbf{26.9} & 28.6 & \textbf{19.7}\\ 
\bottomrule 
\end{tabular}
\caption{Comparison of our method and baselines on the SceneFun3D dataset. All metrics are positively correlated with performance (higher is better).}
\label{table1}
\end{table}

%\begin{table}[t]
%\centering
%\caption{Comparison of our method and baselines on the SceneFun3D dataset. All metrics are positively correlated with performance (higher is better).}
%\setlength{\tabcolsep}{2pt} 
%\footnotesize  
%\begin{tabular}{cccccccc} 
%\toprule 
%{Method} & {$\mathrm{mAP}$} & {$\mathrm{AP}_{50}$} & {$\mathrm{AP}_{25}$} &  {$\mathrm{mAR}$} & {$\mathrm{AR}_{50}$} & {$\mathrm{AR}_{25}$}  & {$\mathrm{mIoU}$}\\ 
%\midrule 
%OpenMask3D & 0.20 & 0.20 & 0.40 & 20.30 & 24.50 & 27.00 & 0.20 \\ 
%LERF & 0.00 & 0.00 & 0.00 & 34.20 & 35.10 & 36.00 & 0.00 \\ 
%Fun3DU & 7.6 & 16.9 & 33.3 & 15.2 \\
%\midrule 
%Ours & \textbf{23.17} & \textbf{26.86} & \textbf{28.61} & \textbf{32.10} & %\textbf{44.19} & \textbf{52.33}& \textbf{19.70}\\ 
%\bottomrule 
%\end{tabular}
%\label{table1}
%\end{table}

\begin{table}[t]
\centering
\begin{tabular}{cccc} 
\toprule 
{Backbone} & {\makecell{Model \\ FLOPS}} & {\makecell{Inference \\ Time (s/sample)}\(\boldsymbol{\downarrow}\)} & {Speed Up\(\boldsymbol{\uparrow}\)}  \\ 
\midrule 
Fun3DU & 48.05G  & 130.26  &  1.00$\times$ \\ 
Ours & 29.50G & \textbf{37.61} & \textbf{3.37}$\times$ \\ 
\bottomrule 
\end{tabular}
\caption{Efficiency comparison between our method and Fun3DU on the SceneFun3D dataset. Inference time per task is reported using an NVIDIA A800 GPU.}
\label{table2}
\end{table}

\section{Experiment}
\subsection{Experimental Settings}
\subsubsection{Dataset.} Evaluation conducted on SceneFun3D~\cite{delitzas2024scenefun3d}, the only existing dataset that provides annotations for 3D functional affordance segmentation in indoor environments. The dataset includes high-resolution 3D scans of real-world indoor scenes and is divided into two subsets: Split0, containing 30 scenes, and Split1, comprising 200 scenes. Each scene is accompanied by approximately 100 high-resolution RGB-D images, along with their corresponding intrinsic and extrinsic camera parameters. In addition, each scene is annotated with about 15 natural language task descriptions and their corresponding ground-truth 3D affordance masks.
We report results on both splits to evaluate performance on small- and large-scale settings.

\subsubsection{Evaluation Metrics.}
We evaluated our model using the standard metrics provided by SceneFun3D~\cite{delitzas2024scenefun3d}, which measure precision in 3D affordance segmentation. Specifically, we reported: 1) Average Precision (AP) at IoU thresholds of 0.25 and 0.5. 2) Mean Average Precision (\(\text{mAP}\)), computed by averaging AP across IoU thresholds from 0.5 to 0.95 in 0.05 increments; 3) Mean Intersection over Union (\(\text{mIoU}\)), which measures per-point segmentation quality across the entire point cloud.

\subsubsection{Baselines.}
We compared our approach against recent open-vocabulary 3D segmentation methods used in the SceneFun3D benchmark. Specifically, we included OpenMask3D~\cite{takmaz2023openmask3d}, LERF~\cite{kerr2023lerf}, OpenIns3D~\cite{huang2024openins3d}, as well as the more recent Fun3DU~\cite{Functionality}, which directly targets 3D scene affordance segmentation. 
%For all baselines, we use the original task description $T$ as input, as it consistently outperforms alternative paraphrased prompts in our preliminary testing.

\subsubsection{Implementation Details.}
We used Qwen~\cite{bai2023qwen} for both affordance concept extraction and the double-check mechanism. For coarse 2D mask generation, we combined MolMo~\cite{deitke2025molmo} with the Segment Anything Model (SAM)~\cite{kirillov2023segment}, leveraging their complementary strengths in vision-language alignment and generic segmentation. The 3D affordance refinement module is implemented using a Point Transformer~\cite{zhao2021point}, trained with the multi-objective loss described in Eqn.~(\ref{loss}). We set the loss weights to: $\alpha_1=0.3$, $\alpha_2=0.3$, $\alpha_3=0.2$, and $\alpha_4=0.2$. For geometric feature extraction, the number of N-nearest neighbors is set to 8,192.

\begin{table}[t]
\centering
%\setlength{\tabcolsep}{1pt} 
%\footnotesize  
\begin{tabular}{ccccccccc} 
\toprule 
\makecell{Affordance\\Weight} & \makecell{Double\\Check} & {$\text{mAP}$} & {$\text{AP}_{50}$} & {$\text{AP}_{25}$} & {$\text{mIoU}$}\\ 
\midrule 
  &  & 3.9 & 6.00 & 14.9 & 4.4 \\ 
 $\checkmark$ &    & 9.3 & 14.2 & 23.3 & 11.3 \\ 
  & $\checkmark$  & 13.1 & 18.1 & 27.9 & 8.5  \\
 $\checkmark$ & $\checkmark$ & \textbf{23.2} & \textbf{26.9} & \textbf{28.6} & \textbf{19.7}\\ 
\bottomrule 
\end{tabular}
\caption{Ablation of the Task-Aware 2D Affordance Detection module on SceneFun3D Split0.}
\label{table3}
\end{table}

\begin{table}[t]
\centering
\begin{tabular}{ccccccc} 
\toprule 
geometry & color & text & {$\text{mAP}$} & {$\text{AP}_{50}$} & {$\text{AP}_{25}$}& {$\text{mIoU}$} \\ 
\midrule 
             &              &  & 7.4 & 18.7 & \textbf{29.9} & 16.7\\ 
$\checkmark$ & $\checkmark$ &  & 19.3 & 19.3 & 24.3 & 15.2\\ 
$\checkmark$ &  & $\checkmark$ & 9.2 & 10.0 & 10.1 & 4.8\\
$\checkmark$ & $\checkmark$ & $\checkmark$ & 5.0 & 5.8 & 6.1 &2.8\\
$\checkmark$ &  &  & \textbf{23.2} & \textbf{26.9} & 28.6 & \textbf{19.7}\\ 
\bottomrule 
\end{tabular}
\caption{Ablation of the 3D Affordance Refinement Module with different input modalities: geometry, color, and text.}
\label{table4}
\end{table}

\begin{figure}[t]
\centering
\includegraphics[width=1\columnwidth]{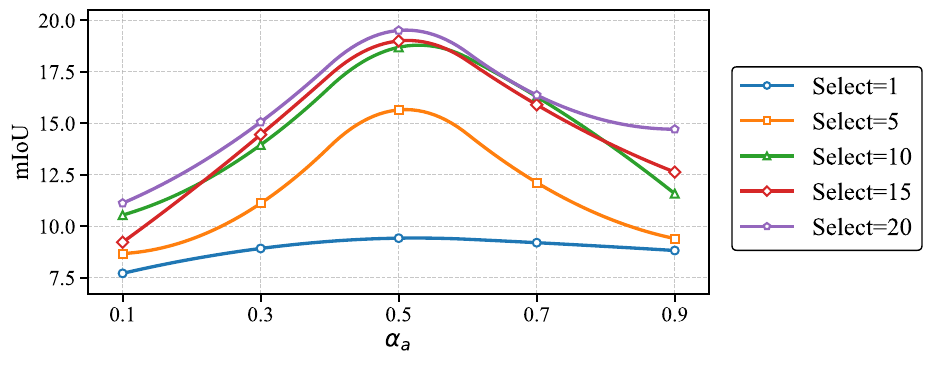} % Reduce the figure size so that it is slightly narrower than the column. Don't use precise values for figure width.This setup will avoid overfull boxes.
\caption{Effect of the number of selected images \(K\)
on segmentation performance, evaluated using \(\text{mIoU}\). Each experiment is conducted with a different Affordance Weight \(\alpha_a\).}
\label{fig_plot}
\end{figure}

\subsection{Quantitative Results}

Table~\ref{table1} presents a comparison of our method with baselines on the SceneFun3D dataset. The results reveal that LERF and OpenIns3D achieve (\text{AP}{25}) scores close to zero, while OpenMask3D attains a modest (\text{AP}{25}) of 0.4.
% Table~\ref{table1} compares our method with baselines on SceneFun3D. LERF's and OpenIns3D's \(\text{AP}_{25}\) hovers near zero while OpenMask3D records only 0.4 in \(\text{AP}_{25}\). 
These limitations primarily arise from undersegmentation issues, such as misclassifying entire cabinets as handles, due to insufficient fine-grained semantic reasoning. These methods struggle with high-resolution point clouds, containing up to 13 million points against only a few hundred annotated ground-truth affordance masks, severely limiting their learning and resulting in low \(\text{AP}\) scores. 
Fun3DU achieves a notable $\text{AP}_{25}$ score of 33.3, likely due to its ability to generate masks that satisfy the lenient $\text{mIoU}$ threshold despite their imprecision. However, its performance is limited by directly translating 2D results into 3D without leveraging geometric structures, leading to a lower $\text{AP}_{50}$ of 16.9 and an $\text{mIoU}$ of 15.2.

In contrast, our method demonstrates a significant advancement in precision, achieving the $\text{AP}_{50}$ of 26.9 and $\text{mIoU}$ of 19.7, surpassing all baselines. This improvement is driven by our synergistic pipeline: 1) CLIP-driven, task-aware frame selection effectively filters irrelevant frames, thereby reducing noise and mitigating undersegmentation issues observed in LERF and OpenMask3D, leading to enhanced (\text{AP}) scores. 2) A coarse-to-fine integration of 2D semantic guidance with 3D geometric reasoning provides richer contextual understanding, resulting in superior (\text{AP}) and (\text{mIoU}) scores compared to Fun3DU.

% In contrast, our method achieves a marked leap in precision as \(\text{AP}_{50}\) of 26.9 and \(\text{mIoU}\) of 19.7 surpass all baselines, demonstrating robust affordance localization even amid texture ambiguity. This stems from our synergistic pipeline: 1) CLIP-driven task-aware frame selection filters out irrelevant frames by combining scene context and affordance relevance effectively reducing noise and addressing undersegmentation issues found in LERF and OpenMask3D, thus improving \(\text{AP}\) scores. 2) Coarse-to-fine integration of 2D semantic guidance and 3D geometric reasoning, which provides richer contextual understanding and leads to stronger \(\text{AP}\)
% and \(\text{mIoU}\) scores than Fun3DU.

Moreover, Table~\ref{table2} quantifies the efficiency of our method compared to Fun3DU, achieving a 3.37× speedup with 40\% fewer FLOPs. This efficiency stems from CLIP-driven task-aware frame selection, which eliminates irrelevant frames and redundant computations, enhancing inference throughput while maintaining performance. This efficient approach is more suitable for embodied downstream applications.

%As shown in Table\ref{table1}, our method outperforms the Fun3DU method across all metrics, which stems from two key innovations enabling superior fine-grained semantic parsing. First, our CLIP-based video frame selection strategy boosts frame-task relevance and computational efficiency. This is reflected in the higher \(\text{AP}\) scores of our method. As shown in Table\ref{table2} this strategy allows for faster processing while achieving better results, and speed measurements comprehensively demonstrate such performance advantages. Second, by leveraging 2D information as guidance and integrating it with 3D geometric cues, our method gains a more comprehensive scene structure understanding. This helps maitain competitive \(\text{AR}\) scores. These results indicate that baseline methods are susceptible to the descriptions of the tasks themselves, focusing more on object-level entities while neglecting specific functional components. The Fun3DU method, on the other hand, requires processing all video frames per inference, leading to extremely long reasoning times, making such approaches hardly suitable for embodied applications. In contrast, our method’s synergistic integration of task-relevant frame selection, along with the fusion of 2D semantic cues and 3D geometric features, not only enhances the precision-recall balance but also establishes a more efficient and effective framework for functional object segmentation, allowing it to outperform Fun3DU in key metrics.

\begin{figure*}[t]
\centering
\includegraphics[width=\textwidth]{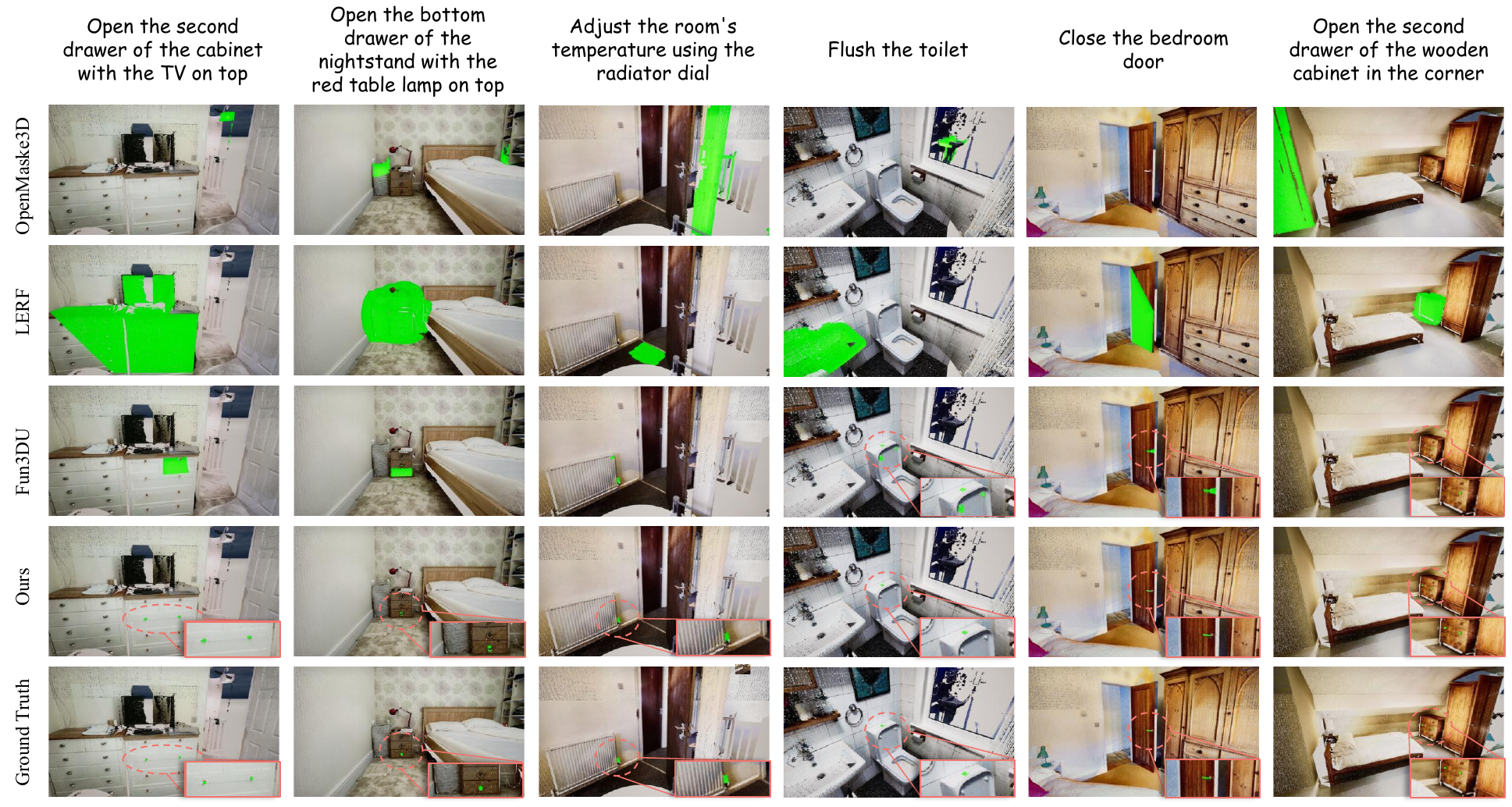} % Reduce the figure size so that it is slightly narrower than the column. Don't use precise values for figure width.This setup will avoid overfull boxes.
\caption{Qualitative comparison on SceneFun3D Split0. Point clouds are cropped around functional objects for improved visibility.}
\label{fig5}
\end{figure*}

\subsection{Ablation Study}
\subsubsection{On Task-Aware 2D Affordance Detection.} 
We evaluated the Affordance-Weighted Frame Selection and Double-Check Point Validation components through ablation experiments as shown in Table~\ref{table3}. The model performs poorly without either component, while enabling each module individually yields moderate improvements. Specifically, the frame selection module raises \(\text{mIoU}\) to 11.3, while the Double-Check mechanism increases \(\text{mAP}\) to 13.1. Notably, activating both modules simultaneously yields the most significant gains across all metrics. These results underscore the complementary roles of the two components: the frame selection module enhances spatial priors by focusing on task-relevant views, while the Double-Check mechanism improves point-level accuracy through geometric validation. Together, they improve segmentation precision and robustness.

Figure~\ref{fig_plot} illustrates the impact of the number of selected $K$-frames on \(\text{mIoU}\). When only one highest-confidence frame is selected, performance is limited due to insufficient multi-view context. Although $K$=20 has the highest \(\text{mIoU}\), 
$K$=10 achieves comparable performance while effectively capturing critical visual cues. Therefore, we select $K$=10
 as it offers a favorable trade-off between accuracy and computational efficiency.

\subsubsection{On 3D Affordance Refinement.}
We conducted ablation experiments reported in Table~\ref{table4}. The results confirm that incorporating geometric structure is critical, yielding the highest scores in \(\text{mAP}\), \(\text{AP}_{50}\), and competitive performance in \(\text{AP}_{25}\). 
We further explore the effect of adding color and text features to the 3D module. Surprisingly, performance degrades across all metrics when these modalities are included. We hypothesize that this decline results from overfitting or confusion introduced by low-quality or sparse semantic signals, especially under the limited data regime of SceneFun3D. These additional modalities may act as noise, hindering convergence and diluting the contribution of reliable geometric cues. This finding highlights the importance of modality alignment: adding more input channels does not guarantee better performance, especially when data is limited. Instead, carefully balancing semantic and geometric information is essential for effective affordance segmentation.  

\subsection{Qualitative Results}
As shown in Figure~\ref{fig5}, %our method produces significantly more accurate and robust affordance segmentation results compared to existing baselines, particularly in complex and cluttered scenes. These improvements stem from the combination of 2D-guided localization and 3D geometry-aware refinement.
OpenMask3D often fails to interpret large-scale point clouds and struggles with basic scene understanding. For example, in the sixth column, it misclassifies upright wall features as part of a cabinet. In many cases, it is unable to localize the target object at all, likely due to its limited semantic grounding and computational inefficiency on dense inputs. LERF focuses on high-level object categories but lacks the ability to isolate task-relevant functional components. In the first and second columns, where the task involves ``opening a cabinet drawer'', LERF highlights the cabinet itself while overlooking the drawer handle, illustrating its tendency to prioritize object names over actionable parts. Fun3DU performs better in identifying functional regions but suffers from spurious mask generation. In the first and second columns, it produces large, imprecise masks due to over-reliance on noisy 2D point predictions from VLMs. Additionally, when nearby textures are visually similar to the affordance region, as in the third and fourth columns, Fun3DU frequently oversegments irrelevant areas. It also struggles with tasks involving multiple disconnected components (e.g., dual handles), as seen in the sixth column.

In contrast, our method produces significantly more accurate and robust affordance segmentation results compared to existing baselines, particularly in complex and cluttered scenes. These improvements stem from the combination of 2D-guided localization and 3D geometry-aware refinement.
%our method benefits from a two-stage design: 1) VLM-guided spatial reasoning provides coarse but task-relevant localization, focusing attention on functional regions; 2) 3D geometry refinement enhances segmentation precision, especially for small, occluded, or texture-ambiguous components. This pipeline enables accurate segmentation of fine-grained interactive parts while effectively suppressing false positives. As demonstrated in Figure~\ref{fig5}, our method delivers task-consistent, spatially precise, and noise-resilient results across diverse scenarios, explaining its superior quantitative performance in Table~\ref{table1}.

\section{Conclusion}
We present a Task-Aware 3D Scene-level Affordance segmentation framework (TASA), that jointly leverages 2D semantic cues and 3D geometric reasoning in a coarse-to-fine manner, marking the first geometry-optimized method for holistic affordance understanding in complex 3D scenes. The framework demonstrates strong performance to high-fidelity point clouds and consistently outperforms baselines in cluttered, real-world environments.

\section{Acknowledgments}
This research was funded by the Zhongguancun Academy (Grant No. 20240306), the National Natural Science Foundation of China (Grant No. 62376140, No. U23A20315 and No. 62376266), the Special Fund for Taishan Scholar Project of Shandong Province.

\bibliography{aaai2026}

\begin{thebibliography}{37}
\providecommand{\natexlab}[1]{#1}

\bibitem[{Achlioptas et~al.(2020)Achlioptas, Abdelreheem, Xia, Elhoseiny, and Guibas}]{achlioptas2020referit3d}
Achlioptas, P.; Abdelreheem, A.; Xia, F.; Elhoseiny, M.; and Guibas, L. 2020.
\newblock Referit3d: Neural listeners for fine-grained 3d object identification in real-world scenes.
\newblock In \emph{European conference on computer vision}, 422--440. Springer.

\bibitem[{Alayrac et~al.(2022)Alayrac, Donahue, Luc, Miech, Barr, Hasson, Lenc, Mensch, Millican, Reynolds et~al.}]{alayrac2022flamingo}
Alayrac, J.-B.; Donahue, J.; Luc, P.; Miech, A.; Barr, I.; Hasson, Y.; Lenc, K.; Mensch, A.; Millican, K.; Reynolds, M.; et~al. 2022.
\newblock Flamingo: a visual language model for few-shot learning.
\newblock \emph{Advances in neural information processing systems}, 35: 23716--23736.

\bibitem[{Bai et~al.(2023)Bai, Bai, Chu, Cui, Dang, Deng, Fan, Ge, Han, Huang et~al.}]{bai2023qwen}
Bai, J.; Bai, S.; Chu, Y.; Cui, Z.; Dang, K.; Deng, X.; Fan, Y.; Ge, W.; Han, Y.; Huang, F.; et~al. 2023.
\newblock Qwen technical report.
\newblock \emph{arXiv preprint arXiv:2309.16609}.

\bibitem[{Chen et~al.(2024)Chen, Xu, Kirmani, Ichter, Sadigh, Guibas, and Xia}]{chen2024spatialvlm}
Chen, B.; Xu, Z.; Kirmani, S.; Ichter, B.; Sadigh, D.; Guibas, L.; and Xia, F. 2024.
\newblock Spatialvlm: Endowing vision-language models with spatial reasoning capabilities.
\newblock In \emph{Proceedings of the IEEE/CVF Conference on Computer Vision and Pattern Recognition}, 14455--14465.

\bibitem[{Chen, Chang, and Nie{\ss}ner(2020)}]{chen2020scanrefer}
Chen, D.~Z.; Chang, A.~X.; and Nie{\ss}ner, M. 2020.
\newblock Scanrefer: 3d object localization in rgb-d scans using natural language.
\newblock In \emph{European conference on computer vision}, 202--221. Springer.

\bibitem[{Chu et~al.(2025)Chu, Deng, Lv, Chen, Li, Hao, and Nie}]{3d-affordancellm}
Chu, H.; Deng, X.; Lv, Q.; Chen, X.; Li, Y.; Hao, J.; and Nie, L. 2025.
\newblock 3d-affordancellm: Harnessing large language models for open-vocabulary affordance detection in 3d worlds.
\newblock \emph{arXiv preprint arXiv:2502.20041}.

\bibitem[{Corsetti et~al.(2025)Corsetti, Giuliari, Fasoli, Boscaini, and Poiesi}]{Functionality}
Corsetti, J.; Giuliari, F.; Fasoli, A.; Boscaini, D.; and Poiesi, F. 2025.
\newblock Functionality understanding and segmentation in 3D scenes.
\newblock In \emph{Proceedings of the Computer Vision and Pattern Recognition Conference}, 24550--24559.

\bibitem[{Deitke et~al.(2025)Deitke, Clark, Lee, Tripathi, Yang, Park, Salehi, Muennighoff, Lo, Soldaini et~al.}]{deitke2025molmo}
Deitke, M.; Clark, C.; Lee, S.; Tripathi, R.; Yang, Y.; Park, J.~S.; Salehi, M.; Muennighoff, N.; Lo, K.; Soldaini, L.; et~al. 2025.
\newblock Molmo and pixmo: Open weights and open data for state-of-the-art vision-language models.
\newblock In \emph{Proceedings of the Computer Vision and Pattern Recognition Conference}, 91--104.

\bibitem[{Delitzas et~al.(2024)Delitzas, Takmaz, Tombari, Sumner, Pollefeys, and Engelmann}]{delitzas2024scenefun3d}
Delitzas, A.; Takmaz, A.; Tombari, F.; Sumner, R.; Pollefeys, M.; and Engelmann, F. 2024.
\newblock Scenefun3d: Fine-grained functionality and affordance understanding in 3d scenes.
\newblock In \emph{Proceedings of the IEEE/CVF Conference on Computer Vision and Pattern Recognition}, 14531--14542.

\bibitem[{Engelmann et~al.(2020)Engelmann, Bokeloh, Fathi, Leibe, and Nie{\ss}ner}]{engelmann20203d}
Engelmann, F.; Bokeloh, M.; Fathi, A.; Leibe, B.; and Nie{\ss}ner, M. 2020.
\newblock 3d-mpa: Multi-proposal aggregation for 3d semantic instance segmentation.
\newblock In \emph{Proceedings of the IEEE/CVF conference on computer vision and pattern recognition}, 9031--9040.

\bibitem[{Gibson(2014)}]{gibson2014ecological}
Gibson, J.~J. 2014.
\newblock \emph{The ecological approach to visual perception: classic edition}.
\newblock Psychology press.

\bibitem[{Huang et~al.(2024)Huang, Wu, Chen, Zhao, Zhu, and Lasenby}]{huang2024openins3d}
Huang, Z.; Wu, X.; Chen, X.; Zhao, H.; Zhu, L.; and Lasenby, J. 2024.
\newblock Openins3d: Snap and lookup for 3d open-vocabulary instance segmentation.
\newblock In \emph{European Conference on Computer Vision}, 169--185. Springer.

\bibitem[{Kerr et~al.(2023)Kerr, Kim, Goldberg, Kanazawa, and Tancik}]{kerr2023lerf}
Kerr, J.; Kim, C.~M.; Goldberg, K.; Kanazawa, A.; and Tancik, M. 2023.
\newblock Lerf: Language embedded radiance fields.
\newblock In \emph{Proceedings of the IEEE/CVF international conference on computer vision}, 19729--19739.

\bibitem[{Kirillov et~al.(2023)Kirillov, Mintun, Ravi, Mao, Rolland, Gustafson, Xiao, Whitehead, Berg, Lo et~al.}]{kirillov2023segment}
Kirillov, A.; Mintun, E.; Ravi, N.; Mao, H.; Rolland, C.; Gustafson, L.; Xiao, T.; Whitehead, S.; Berg, A.~C.; Lo, W.-Y.; et~al. 2023.
\newblock Segment anything.
\newblock In \emph{Proceedings of the IEEE/CVF international conference on computer vision}, 4015--4026.

\bibitem[{Li et~al.(2024)Li, Sun, Sevilla-Lara, and Jampani}]{li2024one}
Li, G.; Sun, D.; Sevilla-Lara, L.; and Jampani, V. 2024.
\newblock One-shot open affordance learning with foundation models.
\newblock In \emph{Proceedings of the IEEE/CVF Conference on Computer Vision and Pattern Recognition}, 3086--3096.

\bibitem[{Li et~al.(2022)Li, Li, Xiong, and Hoi}]{li2022blip}
Li, J.; Li, D.; Xiong, C.; and Hoi, S. 2022.
\newblock Blip: Bootstrapping language-image pre-training for unified vision-language understanding and generation.
\newblock In \emph{International conference on machine learning}, 12888--12900. PMLR.

\bibitem[{Liu et~al.(2023)Liu, Li, Wu, and Lee}]{liu2023visual}
Liu, H.; Li, C.; Wu, Q.; and Lee, Y.~J. 2023.
\newblock Visual instruction tuning.
\newblock \emph{Advances in neural information processing systems}, 36: 34892--34916.

\bibitem[{Liu et~al.(2018{\natexlab{a}})Liu, Wang, Nie, He, Chen, and Chua}]{liu2018attentive}
Liu, M.; Wang, X.; Nie, L.; He, X.; Chen, B.; and Chua, T.-S. 2018{\natexlab{a}}.
\newblock Attentive moment retrieval in videos.
\newblock In \emph{The 41st international ACM SIGIR conference on research \& development in information retrieval}, 15--24.

\bibitem[{Liu et~al.(2018{\natexlab{b}})Liu, Wang, Nie, Tian, Chen, and Chua}]{liu2018cross}
Liu, M.; Wang, X.; Nie, L.; Tian, Q.; Chen, B.; and Chua, T.-S. 2018{\natexlab{b}}.
\newblock Cross-modal moment localization in videos.
\newblock In \emph{Proceedings of the 26th ACM international conference on Multimedia}, 843--851.

\bibitem[{Lu et~al.(2025)Lu, Kong, Huang, and Lee}]{lu2025geal}
Lu, D.; Kong, L.; Huang, T.; and Lee, G.~H. 2025.
\newblock Geal: Generalizable 3d affordance learning with cross-modal consistency.
\newblock In \emph{Proceedings of the Computer Vision and Pattern Recognition Conference}, 1680--1690.

\bibitem[{Luo et~al.(2022)Luo, Fu, Kong, Gao, Ren, Shen, Xia, and Liu}]{luo20223d}
Luo, J.; Fu, J.; Kong, X.; Gao, C.; Ren, H.; Shen, H.; Xia, H.; and Liu, S. 2022.
\newblock 3d-sps: Single-stage 3d visual grounding via referred point progressive selection.
\newblock In \emph{Proceedings of the IEEE/CVF Conference on Computer Vision and Pattern Recognition}, 16454--16463.

\bibitem[{Nguyen et~al.(2023)Nguyen, Vu, Vuong, Nguyen, Vo, Le, and Nguyen}]{nguyen2023open}
Nguyen, T.; Vu, M.~N.; Vuong, A.; Nguyen, D.; Vo, T.; Le, N.; and Nguyen, A. 2023.
\newblock Open-vocabulary affordance detection in 3d point clouds.
\newblock In \emph{2023 IEEE/RSJ International Conference on Intelligent Robots and Systems (IROS)}, 5692--5698. IEEE.

\bibitem[{Qian et~al.(2024)Qian, Chen, Bai, Zhou, Tu, and Li}]{qian2024affordancellm}
Qian, S.; Chen, W.; Bai, M.; Zhou, X.; Tu, Z.; and Li, L.~E. 2024.
\newblock Affordancellm: Grounding affordance from vision language models.
\newblock In \emph{Proceedings of the IEEE/CVF Conference on Computer Vision and Pattern Recognition}, 7587--7597.

\bibitem[{Radford et~al.(2021)Radford, Kim, Hallacy, Ramesh, Goh, Agarwal, Sastry, Askell, Mishkin, Clark et~al.}]{c:1}
Radford, A.; Kim, J.~W.; Hallacy, C.; Ramesh, A.; Goh, G.; Agarwal, S.; Sastry, G.; Askell, A.; Mishkin, P.; Clark, J.; et~al. 2021.
\newblock Learning transferable visual models from natural language supervision.
\newblock In \emph{International conference on machine learning}, 8748--8763. PmLR.

\bibitem[{Shao et~al.(2025)Shao, Zhai, Yang, Luo, Cao, and Zha}]{shao2025great}
Shao, Y.; Zhai, W.; Yang, Y.; Luo, H.; Cao, Y.; and Zha, Z.-J. 2025.
\newblock Great: Geometry-intention collaborative inference for open-vocabulary 3d object affordance grounding.
\newblock In \emph{Proceedings of the Computer Vision and Pattern Recognition Conference}, 17326--17336.

\bibitem[{Sun et~al.(2023)Sun, Qing, Tan, and Xu}]{sun2023superpoint}
Sun, J.; Qing, C.; Tan, J.; and Xu, X. 2023.
\newblock Superpoint transformer for 3d scene instance segmentation.
\newblock In \emph{Proceedings of the AAAI Conference on Artificial Intelligence}, volume~37, 2393--2401.

\bibitem[{Takmaz et~al.(2023)Takmaz, Fedele, Sumner, Pollefeys, Tombari, and Engelmann}]{takmaz2023openmask3d}
Takmaz, A.; Fedele, E.; Sumner, R.~W.; Pollefeys, M.; Tombari, F.; and Engelmann, F. 2023.
\newblock Openmask3d: Open-vocabulary 3d instance segmentation.
\newblock \emph{arXiv preprint arXiv:2306.13631}.

\bibitem[{Van~Vo et~al.(2024)Van~Vo, Vu, Huang, Nguyen, Le, Vo, and Nguyen}]{van2024open}
Van~Vo, T.; Vu, M.~N.; Huang, B.; Nguyen, T.; Le, N.; Vo, T.; and Nguyen, A. 2024.
\newblock Open-vocabulary affordance detection using knowledge distillation and text-point correlation.
\newblock In \emph{2024 IEEE International Conference on Robotics and Automation (ICRA)}, 13968--13975. IEEE.

\bibitem[{Wu et~al.(2024)Wu, Ma, Chen, Wang, Luo, Ji, and Sun}]{wu20243d}
Wu, C.; Ma, Y.; Chen, Q.; Wang, H.; Luo, G.; Ji, J.; and Sun, X. 2024.
\newblock 3d-stmn: Dependency-driven superpoint-text matching network for end-to-end 3d referring expression segmentation.
\newblock In \emph{Proceedings of the AAAI Conference on Artificial Intelligence}, volume~38, 5940--5948.

\bibitem[{Xu et~al.(2024)Xu, Gao, Song, and Hao}]{xu2024weakly}
Xu, L.; Gao, Y.; Song, W.; and Hao, A. 2024.
\newblock Weakly supervised multimodal affordance grounding for egocentric images.
\newblock In \emph{Proceedings of the AAAI Conference on Artificial Intelligence}, volume~38, 6324--6332.

\bibitem[{Xu and Mu(2025)}]{xu2025weakly}
Xu, P.; and Mu, Y. 2025.
\newblock Weakly-supervised affordance grounding guided by part-level semantic priors.
\newblock \emph{arXiv preprint arXiv:2505.24103}.

\bibitem[{Yang et~al.(2023)Yang, Zhai, Luo, Cao, Luo, and Zha}]{grounding3d}
Yang, Y.; Zhai, W.; Luo, H.; Cao, Y.; Luo, J.; and Zha, Z.-J. 2023.
\newblock Grounding 3d object affordance from 2d interactions in images.
\newblock In \emph{Proceedings of the IEEE/CVF International Conference on Computer Vision}, 10905--10915.

\bibitem[{Yang et~al.(2024)Yang, Zhai, Wang, Yu, Cao, and Zha}]{yang2024egochoir}
Yang, Y.; Zhai, W.; Wang, C.; Yu, C.; Cao, Y.; and Zha, Z.-J. 2024.
\newblock Egochoir: Capturing 3d human-object interaction regions from egocentric views.
\newblock \emph{Advances in Neural Information Processing Systems}, 37: 54529--54557.

\bibitem[{Ye and Yu(2024)}]{ye2024pose}
Ye, Q.; and Yu, Z. 2024.
\newblock Pose-Promote: Progressive Visual Perception for Activities of Daily Living.
\newblock \emph{IEEE Signal Processing Letters}.

\bibitem[{Ye et~al.(2025)Ye, Yu, Shao, Cui, Kang, Liu, Torr, and Cao}]{ye2025cat+}
Ye, Q.; Yu, Z.; Shao, R.; Cui, Y.; Kang, X.; Liu, X.; Torr, P.; and Cao, X. 2025.
\newblock Cat+: Investigating and enhancing audio-visual understanding in large language models.
\newblock \emph{IEEE Transactions on Pattern Analysis and Machine Intelligence}.

\bibitem[{Ye et~al.(2024)Ye, Yu, Shao, Xie, Torr, and Cao}]{ye2024cat}
Ye, Q.; Yu, Z.; Shao, R.; Xie, X.; Torr, P.; and Cao, X. 2024.
\newblock Cat: Enhancing multimodal large language model to answer questions in dynamic audio-visual scenarios.
\newblock In \emph{European Conference on Computer Vision}, 146--164. Springer.

\bibitem[{Zhao et~al.(2021)Zhao, Jiang, Jia, Torr, and Koltun}]{zhao2021point}
Zhao, H.; Jiang, L.; Jia, J.; Torr, P.~H.; and Koltun, V. 2021.
\newblock Point transformer.
\newblock In \emph{Proceedings of the IEEE/CVF international conference on computer vision}, 16259--16268.

\end{thebibliography}

\end{document}